\documentclass[sigconf]{acmart}

\renewcommand\footnotetextcopyrightpermission[1]{} 

\pdfoutput=1
\usepackage{booktabs} 
\usepackage{subfig}
\usepackage{graphicx}
\usepackage{float}
\usepackage{amssymb}
\usepackage{amsmath}
\usepackage{enumitem}
\usepackage{booktabs}
\usepackage{textcomp}
\usepackage{array,booktabs,multirow}
\usepackage[skip=0pt]{caption}
\usepackage{hyperref}
\usepackage{graphicx}
\usepackage{amsmath}
\usepackage{algorithm}
\usepackage[noend]{algpseudocode}

\DeclareMathOperator*{\argmin}{argmin}

\begin{document}

\copyrightyear{2019} 
\acmYear{2019} 
\setcopyright{acmlicensed}
\acmConference[KDD '19]{The 25th ACM SIGKDD International Conference on Knowledge Discovery \& Data Mining}{August 03--07, 2019}{Anchorage, Alaska USA}
\acmBooktitle{KDD '19: The 25th ACM SIGKDD International Conference on Knowledge Discovery \& Data Mining, August 03--07, 2019, Anchorage, Alaska USA}

\title{Target-wise Kernel Mean Embedding in Multiple Instance Regression}

\author{Thomas Uriot}
\affiliation{%
  \institution{Department of Computing}
  \institution{Imperial College London}
}
\email{tmu15@ic.ac.uk}

\renewcommand{\shortauthors}{Uriot}

\begin{abstract}
In this paper, we propose an extension to an existing algorithm (instance-MIR) which tackles the multiple instance regression (MIR) problem, also known as distribution regression. The MIR setting arises when the data is a collection of bags, where each bag consists of several instances which correspond to the same and unique real-valued label. The goal of a MIR algorithm is to find a mapping from the instances of an unseen bag to its target value. The instance-MIR algorithm treats all the instances separately and maps each instance to a label. The final bag label is then taken as the mean or the median of the predictions for
that given bag. While it is conceptually simple, taking a single statistic to summarize the distribution of the labels in each bag is a limitation. In spite of this performance bottleneck, the instance-MIR algorithm has been shown to be competitive when compared to the current state-of-the-art methods. We address the aforementioned issue by computing the kernel mean embeddings of the distributions of the predicted labels, for each bag, and learn a regressor from these embeddings to the bag label. We test our algorithm (instance-kme-MIR) on five real world datasets and obtain better results than the baseline instance-MIR across all the datasets, while achieving state-of-the-art results on two of the datasets.
\end{abstract}


\maketitle

\section{Introduction}

Multiple instance learning (MIL) is a setting which falls under the supervised learning paradigm. Within the MIL framework, there exist two different learning tasks: multiple instance classification (MIC) \cite{amores2013multiple} and multiple instance regression (MIR) \cite{dietterich1997solving, ray2001multiple}. The former has been extensively studied in the literature while the latter has been underrepresented. This could be due to the fact that many of the data sources studied within the MIL framework are images and text, which correspond to classification tasks. The MIC problem generally consists in classifying bags into positive or negative examples where negative bags contain only negative instances and positive bags contain at least one positive instance. A multitude of applications are covered by the MIC framework. It has been applied to medical imaging in a weakly supervised setting \cite{Jiajun-15, Xu-14} where each image is taken as a bag and sub-regions of the image are instances, to image categorization \cite{chen2004image} and retrieval \cite{zhang2002content, yang2000image} and to analyzing videos \cite{sikka2013weakly}, where the video is treated as the bag and the frames are the instances. 
 
On the other hand, the MIR problem, where bags labels are now real valued, has been much less prevalent in the literature. In a regression setting, as opposed to classification, one cannot simply identify a single positive instance. Instead, one needs to estimate the contribution of each of the instances towards the bag label. The MIR problem was first introduced in the context of predicting drug activity level \cite{dietterich1997solving} and the first proposed MIR algorithm relied on the assumption that the bag's label can be fully explained by a single prime instance (prime-MIR) \cite{ray2001multiple}. However, this is a simplistic assumption as we throw away a lot of information about the distribution (e.g, variance, skewness). Instead of assuming that a single instance is responsible for the bag's label, the MIR problem has been tackled using a weighted linear combination of the instances \cite{wagstaff2007salience}, or as a prime cluster of instances (cluster-MIR) \cite{wagstaff2008multiple}. Other works have looked at first efficiently mapping the instances in each bag to a new embedding space, and then train a regressor on the new embedded feature space. For instance, one can transform the MIR problem to a regular supervised learning problem by mapping each bag to a feature space which is characterized by a similarity measure between a bag and an instance \cite{chen2006miles}. The resulting embedding of a bag in the new feature space represents how similar a bag is to various instances from the training set. A drawback of this approach is that the embedding space for each bag can be high-dimensional when the number of instances in the training set is large, producing many redundant and possibly uninformative features.
 
In this paper, we use a similar approach and compute the kernel mean embeddings for each bag \cite{muandet2017kernel}. The use of kernel mean embedding in distribution regression has been applied to various real-world problems such as analyzing the 2016 US presidential election \cite{flaxman2016understanding} and estimating aerosol levels in the atmosphere \cite{szabo2015two}. Intuitively, kernel mean embedding measures how similar each bag is to all the other bags from the training set. In this paper, as opposed to previous works, we do not compute the kernel mean embeddings directly on the input features (i.e, on the instances) but on the predictions made by a previous learning algorithm (e.g, a neural network). This insight comes from the fact that a simple baseline algorithm (instance-MIR) performed surprisingly well on several datasets, when the regressor was a neural network with a large hidden layer \cite{uriot2019learning}. The instance-MIR algorithm essentially ignores the fact that we are in a distribution regression framework and treats each instance as a separate observation, thereby yielding a unique prediction for each instance. The final bag label is taken to be the mean or the median of the predictions for
that given bag. However, using a point estimate in the original prediction space is a performance bottleneck. In this paper, we propose a novel algorithm (instance-kme-MIR), which leverages both the representational power of the instance-MIR algorithm equipped with a neural network and alleviate the aforementioned issue by mapping our predictions into a high or infinite-dimensional space, characterized by a kernel function. We test our approach on 5 remotely sensed real-world datasets.

\section{Related Work}

The datasets we are using to test our algorithm stems from remotely sensed data\footnote{\url{http://www.dabi.temple.edu/~vucetic/MIR.html}} \footnote{\url{https://harvist.jpl.nasa.gov/papers.shtml}}, and have previously been described \cite{uriot2019learning, wang2012mixture} and studied as a distribution regression problem \cite{wang2008aerosol, wang2012mixture, uriot2019learning}. This allows us to compare the performance of our approach with the baseline instance-MIR and the current state-of-the-art. The first application (3 of the 5 datasets) consists in predicting aerosol optical depth (AOD) - aerosols are fine airborne solid particles or liquid droplets in air, that both reflect and absorb incoming solar radiation. The second application (2 of the 5 datasets) is the prediction of county-level crop yields \cite{wagstaff2007salience} (wheat and corn) in Kansas between 2001 and 2005. These two applications can naturally be framed as a multiple instance regression problem. Indeed, in both applications, satellites will gather noisy measurements due to the intrinsic variability within the sensors and the properties of the targeted area on Earth (e.g, surface and atmospheric effects). For the AOD prediction task, aerosols have been found to have a very small spatial variability over distances up to 100 km \cite{ichoku2002spatio}. For the crop data, we can reasonably assume that the yields are similar across a county and thus consider the bag label as the aggregated yield over the entire county.

The first study which investigated estimating AOD levels within a MIR setting, proposed an iterative method (pruning-MIR) which prunes outlying instances from each bag and then proceeds in a similar fashion as instance-MIR \cite{wang2008aerosol}. 
The main drawback of this approach is that it is not obvious what the pruning threshold should be and we may thus get rid of informative instances in the process. In a subsequent work, the authors investigated a probabilistic framework (EM-MIR) by fitting a mixture model and using the expectation-maximization (EM) algorithm to learn the mixing and distribution parameters \cite{wang2012mixture}. The current state-of-the-art algorithm (attention-MIR) on the AOD datasets has been obtained by treating each bag as a set (i.e, an unordered sequence) of instances \cite{uriot2019learning}. To do so, the authors implemented an order-invariant operation characterized by a content-based attention mechanism \cite{vinyals2015order}, which then attends the instances a selected number of times. Finally, the problem of estimating AOD levels has been tackled using kernel mean embedding directly on the input features (i.e, the instances) \cite{szabo2015two}, where they show that performance is robust to the kernel choice but the hyperparameter values of the kernels are of primary importance. 
In this paper, however, we compute the kernel mean embeddings of the distributions of the predicted labels made by a neural network. 
In order to have a principled way to find the kernel parameters, authors have proposed a Bayesian kernel mean embedding model with a Gaussian process prior, from which we can obtain a closed form marginal pseudolikelihood \cite{flaxman2016bayesian}. This marginal likelihood can then be optimized in order to find the kernel parameters.

\section{Background}

\subsection{Multiple Instance Regression}

In the MIR problem, our observed dataset is $\{(\{x_{i,l}\}_{l=1}^{L_i}, y_i)\}_{i=1}^{B}$, where B is the number of bags, $y_i \in \mathbb{R}$ is the label of bag $i$, $x_{i,l}$ is the $l^{th}$ instance of bag $i$ and $L_i$ is the number of instances in bag $i$. Note that $x_{i,l} \in \mathcal{X}$, and $\mathcal{X}$ is a subset of $\mathbb{R}^{d}$, where $d$ is the number of features in each instance. The number of features must be the same for all the instances, but the number of instances can vary within each bag. 

We want to learn the best mapping $\hat{f}$: $\{x_{i,l}\}_{l=1}^{L_i} \to \hat{y}_i$, $i=1\ldots B$. By best mapping we mean the function $\hat{f}$ which minimizes the mean squared error (MSE) on bags unseen during training (e.g, on the validation set). Formally, we seek $\hat{f}$ such that

\begin{equation}
\hat{f} = \textrm{arg min}_{f \in \mathcal{H}} \hspace{0.1cm} \frac{1}{B^*} \sum_{i=1}^{B^*} MSE(y_{i}^{*},f(\{x_{i,l}^{*}\}_{l=1}^{L_{i}^{*}})),
\end{equation}

from the validation data $\{(\{x_{i,l}^*\}_{l=1}^{L_{i}^*}, y_i^*)\}_{i=1}^{B^*}$, where $\mathcal{H}$ is the hypothesis space of functions $f$ under consideration. 

The two main challenges that the multiple instance regression problem poses are to find which instances are predictive of the bag's label and to efficiently summarize the information from the instances within each bag. However, the instance-MIR baseline algorithm, which we describe next, does not attempt to solve the multiple instance regression problem by addressing the two aforementioned challenges. Instead, it simply treats each instance independently and fit a regression model to all the instances separately. 
\subsection{Instance-MIR Algorithm}

As mentioned, the instance-MIR algorithm makes predictions on all the instances before taking the mean or the median of the predictions for each bag, as the final prediction. This means that during training, all the instances have the same weights and thus contribute equally to the loss function.

Formally, our dataset is formed by pairs of instance and bag label which can be denoted as $\{(x_{i,l}, y_i), \hspace{0.1cm} i=1\ldots B, \hspace{0.1cm} l=1\ldots L_i\}$. 
The final label prediction on an unseen bag can be simply calculated as

\[\hat{y}_i^* = \frac{1}{L_i^*}\sum_{l=1}^{L_i^*}\hat{y}_{i,l}^*, \hspace{0.2cm} i=1\ldots B^*,\]

where $\hat{y}_{i,l}^*$ is the predicted label corresponding to the $l_{th}$ instance in bag $i$. Empirically, this method has been shown to be competitive \cite{ray2005supervised}, even though it requires models with high complexity in order to be able to effectively map many different noisy instances to the same target value. Thus, it is appropriate to take $\hat{f}$ as a neural network with a large number of hidden units \cite{uriot2019learning}, as apposed to a small number \cite{wang2012mixture}.

\subsection{Kernel Mean Embedding}

In this subsection, we briefly describe kernel mean embedding and its application to distribution regression, where the goal is to compute the kernel mean embedding of each bag. We assume that the instances $\{x_{i,l}\}_{l=1}^{L_i}$ in each bag, are i.i.d. samples from some unobserved distribution $P^i$, for $i=1,\ldots,B$. The idea is to adopt a two-stage procedure by first representing each set of samples (i.e, bags) $\{x_{i,l}\}_{l=1}^{L_i}$ by its corresponding kernel mean embedding and then train a kernel ridge regression on those embeddings \cite{szabo2015two}. 

Formally, let $\mathcal{H}_k$ be a reproducing kernel Hilbert space (RKHS), which is a potentially infinite dimensional space of functions $f: \mathcal{X} \to \mathbb{R}$, and let $k: \mathcal{X} \times \mathcal{X} \to \mathbb{R}$ be a reproducing kernel function of $\mathcal{H}_k$. Then for $f \in \mathcal{H}_k, x \in \mathcal{X}$, we can evaluate $f$ at $x$ as an inner product $f(x)=\langle f, k(\cdot,x)\rangle_{\mathcal{H}_k}$ (reproducing kernel property). Then, for a probability measure $\textrm{P} \in \mathcal{X}$ we can define its kernel mean embedding as 

\begin{equation}
\mu_{\textrm{P}} = \int k(\cdot, x)\textrm{P}(dx) \in \mathcal{H}_k.
\end{equation}

For $\mu_P$ to be well-defined, we simply require that the norm of $k$ is finite, and so we want $k(\cdot,x)$ such that $\int \sqrt{k(x, x)}\textrm{P}(dx) < \infty$. This is always true for kernel functions that are bounded (e.g, Gaussian RBF, inverse multiquadric) but may be violated for unbounded ones (e.g, polynomial) \cite{szabo2015two}. In fact, it has been shown that the kernel mean embedding approach to distribution regression does not yield satisfying results when using a polynomial kernel, due to the aforementioned violation \cite{szabo2015two}.

However, as mentioned, we do not have access to $\textrm{P}^i$ but only observe i.i.d. samples $\{x_{i,l}\}_{l=1}^{L_i}$ drawn from it. Instead, we compute the empirical mean estimator $\hat{\mu}_P$ of $\mu_P$, given by 

\begin{equation}
\hat{\mu}_{\textrm{P}_i} = \int k(\cdot, x)\hat{\textrm{P}}^i(dx) = \frac{1}{L_i}\sum_{l=1}^{L_i} k(\cdot,x_{i,l}), \hspace{0.2cm} \textrm{for bag $i$}.
\end{equation}

\subsection{Kernel Ridge Regression}

In kernel ridge regression (KRR), we seek to find the set of parameters $\hat{\alpha}$, such that

\begin{equation}
\hat{\alpha} = \argmin_{\alpha}(\|y-\textbf{K}\alpha\|^2+\lambda \alpha^T\textbf{K}\alpha),
\end{equation}

where $\textbf{K} \in \mathbb{R}^{n \times n}$ is the kernelized Gram matrix of the dataset, and $\lambda$ is the hyperparameter controlling the amount of weight decay (i.e, $L_2$ regularization) on the parameters $\alpha$. In the case of KRR applied to kernel mean embedding, we have 

\begin{equation}
\textbf{K}(i,j) = k^{'}(\hat{\mu}_{\textrm{P}_i},\hat{\mu}_{\textrm{P}_j}), \hspace{0.2cm} \textrm{for bags $i,j = 1,\ldots,B$,}
\end{equation}

where $k^{'}$ is the KRR kernel and $\textbf{K} \in \mathbb{R}^{B \times B}$ ($B$ is the number of bags in the training set). In this paper, we take $k^{'}$ to be the linear kernel, as it simplifies the computation and has been shown to yield competitive results when compared to non-linear kernels \cite{szabo2015two}. Thus, we have that

\begin{equation}
\begin{aligned}
\hspace{-0.2cm}\textbf{K}(i,j) ={} &  k^{'}(\hat{\mu}_{\textrm{P}_i},\hat{\mu}_{\textrm{P}_j}) = \langle \hat{\mu}_{\textrm{P}_i}, \hat{\mu}_{\textrm{P}_j}\rangle_{\mathcal{H}_k} \\
& \hspace{-0.4cm} =  \Big< \frac{1}{L_i}\sum_{l=1}^{L_i} k(\cdot,x_{i,l}), \frac{1}{L_j}\sum_{m=1}^{L_j}  k(\cdot,x_{j,m})\Big>_{\mathcal{H}_k} \\
& \hspace{-0.4cm} = \frac{1}{L_i} \frac{1}{L_j} \sum_{l=1}^{L_i} \sum_{m=1}^{L_j} k(x_{i,l}, x_{j,m}), \hspace{0.2cm}
\end{aligned}
\end{equation}

where $x_{i,l}$ is the $l^{th}$ instance of bag $i$ and $x_{j,m}$ is the $m^{th}$ instance of bag j and the last equality is due to the reproducing property.
In order to make predictions $\hat{y}_{test}$ on bags not seen during training, we simply compute
\begin{equation}
\hat{y}_{test} = \hat{\alpha} \textbf{K}_{test},
\end{equation}

where $\hat{\alpha} = y_{train} (\textbf{K}_{train} + \lambda \textbf{I}_{B \times B})^{-1}$ is obtained by differentiating (4) with respect to $\alpha$, equating to 0, and solving for $\alpha$. Note that as mentioned in subsection 3.1, $B$ is the number of bags in the training set and $B^*$ is the number of unseen bags (e.g, in validation or testing set), and so $\textbf{K}_{test} \in \mathbb{R}^{B \times B^*}$.

\section{Instance-kme-MIR Algorithm}

In this section, we describe our novel algorithm (instance-kme-MIR), and discuss the choice we made for the hyperparameter values. We emphasise that the novelty in this paper is to compute the kernel mean embeddings on the predictions made by a previous learning algorithm, as opposed to previous works \cite{szabo2015two, law2017bayesian, flaxman2016understanding, flaxman2016bayesian}, where the authors directly compute the kernel mean embeddings on the input features. Our algorithm can be seen as an extension of instance-MIR, where we take advantage of the representational power of neural networks (Part 1 of our algorithm), and address its performance bottleneck by computing the kernel mean embeddings on the predictions (Part 2 of our algorithm). 

In our implementation\footnote{ \texttt{https://github.com/pinouche/Instance-kme-MIR}}, we choose $D=50$ and $f$ to be a single layered neural network, as it was shown to yield good results for the instance-MIR algorithm \cite{uriot2019learning}. We purposefully set the number of folds $D$ to be large, so that in Part 1 of our algorithm, we still train $f$ on $98\%$ of the training set. It thus makes sense to use the same hyperparameter values for the neural network $f$ when comparing the baseline instance-MIR to instance-kme-MIR. For Part 2, we experimented with two different kernels $k$ (RBF and inverse multiquadric). 

\section{Evaluation}

\subsection{Training Protocol}

In order to fairly compare our algorithm to the current state-of-the-art \cite{uriot2019learning, wang2012mixture}, we evaluate its performance using the same training and evaluation protocol. The protocol consists in a 5-fold cross validation, where the bags in the training set are randomly split into 5 folds, out of which 4 folds are used in training and 1 fold serves as the validation set. In turn, each of the 5 folds serves as the validation set and the 4 remaining folds as the training set. The cross validation is repeated 10 times in order to eliminate the randomness involved in choosing the folds. We use the root mean squared error (RMSE) to evaluate the performance and report our results, shown in Table 1, on 5 real-world datasets. While the baseline instance-MIR was already evaluated \cite{uriot2019learning}, we re-implement it on the 3 AOD datasets, with different hyperparameter values, and thus obtain distinct results. The validation loss reported in Table 1 below is the average loss over the 50 evaluations (10 iterations of 5-fold cross validation).

\begin{algorithm}[H]
\caption{Instance-kme-MIR Algorithm}

\hspace{-1.9cm}\textbf{Inputs:} $\left(\{(\{x_{i,l}\}_{l=1}^{L_i}, y_i)\}_{i=1}^{B}, \{(\{x_{i,l}^*\}_{l=1}^{L_{i}^*}, y_i^*)\}_{i=1}^{B^*}\right)$ \\
\hspace{-1cm}\textbf{Outputs:} Bag level predictions $\{\hat{y}_i^*\}_{i=1}^{B^*}$ on validation set \\

\hspace{-4.6cm}\textbf{Part 1:} Out-of-fold stacking 
\begin{algorithmic}[1]
\State \textbf{Set} $D$ = \textit{number\_folds}
\State \textbf{Initialize} an array $\textbf{A}$ with $\sum_{i=1}^{B}L_i$ elements (i.e, number of training instances) 
\State \textbf{Choose} a learning algorithm $f$
\State \textbf{Shuffle} the bags in the training data $\{(\{x_{i,l}\}_{l=1}^{L_i}, y_i)\}_{i=1}^{B}$
\State \textbf{Partition} the training data $\{(\{x_{i,l}\}_{l=1}^{L_i}, y_i)\}_{i=1}^{B}$ \textrm{into $D$ folds, with an equal number of bags in each fold:}  $\left\{\{(\{x_{i,l}\}_{l=1}^{L_i}, y_i)\}_{i=F_0}^{F_1},\ldots,\{(\{x_{i,l}\}_{l=1}^{L_i}, y_i)\}_{i=F_{D-1}}^{F_D}\right\}$, \textrm{where} $F_0=1, F_D=B$
\For{\textrm{$k=0,\ldots,D-1$}}
        \State \textbf{Set} \textit{counter} = 0
        \State \textbf{Set} $\{X,Y\}_{\textrm{train}}$ = $\left\{\{(\{x_{i,l}\}_{l=1}^{L_i}, y_i)\}_{i=1}^{B}\right\}_{-k}$ (take all the bags except those in fold $k$) 
        \State \textbf{Set} $\{X,Y\}_{\textrm{val}}$ = $(\{x_{i,l}\}_{l=1}^{L_i}, y_i)\}_{i=F_{k}}^{F_{k+1}}$ (take all the bags in fold $k$) 
        \State \textbf{Learn} $\hat{f}$: $x_{i,l} \to \Hat{y}_{i,l}$, $i=\{1,\ldots, B\}_{-k}$, $l=1,\ldots, L_i$ (see equation (1))
        \For{$i=F_k,\ldots,F_{k+1}$}
        \For{$l=1,\ldots,L_i$}
        \State \textbf{Predict} $\Hat{y}_{i,l} = \Hat{f}(x_{i,l})$,
        \State \textbf{Set} $\textbf{A}[\textit{counter}] = \Hat{y}_{i,l}$ (build a stacked training set for Part 2)
        \State \textit{counter} += 1 \vspace{0.1cm}
        \EndFor{\textbf{end for}} \vspace{0.1cm}
        \EndFor{\textbf{end for}} \vspace{0.1cm}
        \EndFor{\textbf{end for}}
\State \textbf{Return} $\textbf{A}$ 
\end{algorithmic}
\textbf{Part 2:} Kernel mean embedding and KRR on the stacked dataset $\textbf{A}$ 
\begin{algorithmic}[1]
\State \textbf{Choose} the weight decay value $\lambda$
\State \textbf{Choose} the kernel function $k$ (and its parameter values)
\State \textbf{Compute} $\textbf{K}(i,j)_{train} = \frac{1}{L_i} \frac{1}{L_j} \sum_{l=1}^{L_i} \sum_{m=1}^{L_j} k(\Hat{y}_{i,l}, \Hat{y}_{j,m})$, $i,j=1,\ldots,B$ (see equation (6)-(7))
\State \textbf{Compute} $\Hat{\alpha} = y_{train} (\textbf{K}_{train} + \lambda \textbf{I}_{B \times B})^{-1}$, \textrm{where} $y_{train} = [y_1,\ldots,y_B] \in \mathbb{R}^{B}$ 
\State \textbf{Return} $\Hat{\alpha}$
\end{algorithmic}
\textbf{Part 3}: Predict bag labels $\{\hat{y}_i^*\}_{i=1}^{B^*}$ on the validation set
\begin{algorithmic}[1]
\For{$i=1,\ldots,B^*$}
        \For{$l=1,\ldots,L_i^*$}
        \State \textbf{Predict} $\Hat{y}_{i,l}^* = \Hat{f}(x_{i,l}^*)$,
        \EndFor{\textbf{end for}} \vspace{0.1cm}
        \EndFor{\textbf{end for}}
\State \textbf{Compute} $\textbf{K}(i,j)_{val} = \frac{1}{L_i} \frac{1}{L_j^*} \sum_{l=1}^{L_i} \sum_{m=1}^{L_j^*} k(\Hat{y}_{i,l}, \Hat{y}_{j,m}^*)$, $i=1,\ldots,B$, $j=1,\ldots,B^*$
\State \textbf{Compute} $\Hat{y}_{val} = \Hat{\alpha}\textbf{K}_{val}$, \textrm{where} $\Hat{y}_{val} = [\Hat{y}_1^*,\ldots,\Hat{y}_{B*}^*] \in \mathbb{R}^{B^*}$ (see equation (8)) 
\State \textbf{Return} $\Hat{y}_{val}$ (i.e, $\{\hat{y}_i^*\}_{i=1}^{B^*}$)
\end{algorithmic}
\end{algorithm}

\subsection{Results}

In Table 1, we display the results for 4 algorithms: the baseline instance-MIR (described in subsection 3.2), attention-MIR \cite{uriot2019learning}, EM-MIR \cite{wang2012mixture} and our novel algorithm (instance-kme-MIR), for two different kernels $k_{\textrm{RBF}}$ and $k_{\textrm{INV}}$, where

\begin{equation*}
k(x,x^\prime)_{RBF} = \textrm{exp}\left(-\frac{\|x-x^\prime\|^2}{2\theta^2}\right), \hspace{0.2cm} k(x,x^\prime)_{INV} = \frac{1-\|x-x^\prime\|^2}{{\|x-x^\prime\|^2}+\theta}. \end{equation*}

Note that prior to our implementation of instance-kme-MIR, the state-of-the-art results on the 5 datasets were shared between the 3 other algorithms \cite{uriot2019learning}. Now, as can be seen in Table 1, attention-MIR achieves the best results on the AOD datasets while instance-kme-MIR yields the best results on the crop datasets.

We experimented with several values for $\theta$ and $\lambda$, where $\theta \in \{10, 20, \ldots, 130, 140\}$ and $\lambda \in \{10^{-1}, 10^{-2}, \ldots, 10^{-15}, 10^{-16}\}$, with a constant increment for both hyperparameters. The results in Table 1 are reported for the best hyperparameter values. We found that while extreme hyperparameter values negatively impacted the performance of our algorithm, most values yielded similarly good results, which means that our algorithm is robust to hyperparameter values.

\begin{table*}[t]
\centering
\caption{The loss for the 3 AOD datasets (MISR1, MISR2, MODIS) is the RMSE $\times$ 100 and for the 2 CROP datasets (WHEAT, CORN) the loss is the RMSE.} 
\begin{tabular}{c|ccccc}
                                & \multicolumn{5}{c}{\textbf{Datasets}}                                                                                                    \\
\multicolumn{1}{l|}{\hspace{0.7cm} Algorithms} & \multicolumn{1}{l}{MODIS} & \multicolumn{1}{r}{MISR1} & \multicolumn{1}{l}{MISR2} & \multicolumn{1}{l}{WHEAT} & \multicolumn{1}{l}{CORN} \\ \hline
Instance-MIR (mean)                      & 10.4   & 9.02& 7.61 & 4.96                      & 24.57            \\
Instance-MIR (median)                      &   10.4    &   8.89     &  7.50  &5.00 & 24.72           \\
Instance-kme-MIR ($k_{\textrm{INV}}$)   & 10.1& 8.68&7.28  & 4.91     &   \textbf{24.40}        \\
Instance-kme-MIR ($k_{\textrm{RBF}}$)   & 10.1& 8.70&7.38  & \textbf{4.90}     &   24.51        \\
EM-MIR \citep{wang2012mixture}   & 9.5 & 7.5  & 7.3                       & \textbf{4.9}              & 26.8                      \\
Attention-MIR \cite{uriot2019learning}                     & \textbf{9.05}             & \textbf{7.32}             & \textbf{6.95}             & 5.24                      & 27.00                      \\ \bottomrule
\end{tabular}
\label{tab3}
\end{table*}

Instance-MIR (median) refers to the instance-MIR algorithm where the median is used to compute the final prediction for each bag, instead of the mean, as described in subsection 3.2. We can see that there does not seem to be an advantage to using the mean or the median, as both methods achieve very similar results. On the other hand, we can see that our algorithm consistently outperforms the baseline instance-MIR. However, note that since our algorithm makes use of the predictions made from instance-MIR (in Part 2 of Algorithm 1), we can only aim to achieve a measured improvement over the standard instance-MIR. Thus, our method is mostly beneficial in the cases where instance-MIR is the best out-of-the box algorithm (e.g, on the 2 crop datasets). Since our algorithm computes the kernel mean embedding between scalars (i.e, between the real-valued predictions) and is robust to values of $\lambda$ and $\theta$, it is easy to tune and its computation cost is very close to that of instance-MIR.

\section{Conclusion}

In this paper, we developed a straightforward extension of the baseline instance-MIR algorithm. Our method takes advantage of the expressive power of neural networks while addressing the main weakness of instance-MIR by computing the kernel mean embeddings of the predictions. We have shown that our algorithm consistently outperforms the baseline and achieves state-of-the-art results on the 2 crop datasets. In addition, our algorithm is robust to the kernel parameter values and its performance gains come at a low computational cost. 

Nonetheless, it fails when the baseline instance-MIR does not yield satisfying results (e.g, on the 3 crop datasets). This is because we compute the kernel mean embeddings on predictions made from the baseline instance-MIR, and we can thus only expect measured improvements from that baseline. Another drawback of our method comes from the fact that instance-MIR assigns the same weights to all the instances during training. However, the number of instances per bag may vary and it would make sense to be more confident when we make a prediction on a bag which contains a large number of instances compared to a bag with only a few instances. To tackle this issue, we could take a Bayesian approach to kernel mean embedding and explicitly express our uncertainty in the sampling variability of the groups \cite{law2017bayesian}. 

Finally, as future work, we could use the attention coefficients from attention-MIR, in order to weigh the contribution of each of the instances towards the loss function. This would get rid of potentially redundant and noisy instances, thus improving the quality of the training data.


\end{document}